\documentclass[letterpaper, 10 pt, journal, twoside]{IEEEtran}

\IEEEoverridecommandlockouts                              

\usepackage{graphics} 
\usepackage{graphicx}
\usepackage{siunitx}    

\usepackage{amsmath} 
\usepackage{algpseudocode}
\usepackage{algorithm}
\usepackage{cite}
\usepackage{balance}
\usepackage[normalem]{ulem}
\newcommand{\change}[1]{#1}
\newcommand{\delete}[1]{}

\title{\LARGE \bf

Seeing Through Pixel Motion:  Learning Obstacle Avoidance from Optical Flow with One Camera

}

\author{Yu Hu$^1$, Yuang Zhang$^1$, Yunlong Song$^2$, Yang Deng$^1$, Feng Yu$^1$, Linzuo Zhang$^1$ \\ Weiyao Lin$^1$
, Danping Zou$^{1\dag}$, and Wenxian Yu$^1$  

\thanks{Manuscript received: September 19, 2024; Revised January 11, 2025; Accepted March 20, 2025.}
\thanks{This paper was recommended for publication by Editor Aniket Bera. Name upon evaluation of the Associate Editor and Reviewers' comments.}
\thanks{$^{\dag}$ Corresponding author ({\tt dpzou@sjtu.edu.cn}). $^1$Shanghai Jiao Tong University. $^2$Y. Song is independent researcher. This work was supported by National Key R\&D Program of China (2022YFB3903802) and National Science Foundation of China (62073214).} 
\thanks{Digital Object Identifier (DOI): see top of this page.}
}

\begin{document}

\maketitle

\markboth{IEEE Robotics and Automation Letters. Preprint Version. Accepted March, 2025}
{Hu \MakeLowercase{\textit{et al.}}: Seeing Through Pixel Motion: Learning Obstacle Avoidance from Optical Flow with One Camera} 

\begin{abstract}

Optical flow captures the motion of pixels in an image sequence over time, providing information about movement, depth, and environmental structure. Flying insects utilize this information to navigate and avoid obstacles, allowing them to execute highly agile maneuvers even in complex environments. Despite its potential, autonomous flying robots have yet to fully leverage this motion information to achieve comparable levels of agility and robustness.
\change{The main challenges are two-fold: (1) extracting accurate optical flow from visual data during high-speed flight and (2) designing a robust controller that can handle noisy optical flow estimations while ensuring robust performance in complex environments.}
%
To address these challenges, we propose a novel end-to-end system for quadrotor obstacle avoidance using monocular optical flow. We develop an efficient differentiable simulator coupled with a simplified quadrotor model, allowing our policy to be trained directly through first-order gradient optimization. 
%
%
Additionally, we introduce a central flow attention mechanism and an action-guided active sensing strategy that enhances the policy’s focus on task-relevant optical flow observations to enable more responsive decision-making during flight. 
Our system is validated both in simulation and the real world using an FPV racing drone. Despite being trained in a simple environment in simulation, our system demonstrates agile and robust flight in various unknown, cluttered environments in the real world at speeds of up to~\SI{6}{\meter\per\second}.

\end{abstract}

\begin{IEEEkeywords}
Aerial Systems: Perception and Autonomy; Collision Avoidance; Sensorimotor Learning
\end{IEEEkeywords}
\section{INTRODUCTION}

\change{
\IEEEPARstart{A}{} fundamental challenge for micro aerial vehicles (MAVs) is achieving agile and safe navigation in cluttered environments. Multiple planning and control algorithms have been proposed to address this, showcasing impressive progress in swarm navigation \cite{zhou2022swarm, zhou_ego-planner_2020, Zhang2024BackTN} and agile flight in dense forests \cite{loquercio_learning_2021, Zhang2024BackTN, yu2024mavrl}. 
A key component of these approaches is their reliance on depth or stereo cameras, which provide direct measurements of obstacle distances to inform planning and control strategies.
}


\begin{figure}[thbp]
  \centering
  \includegraphics[width=0.48\textwidth]{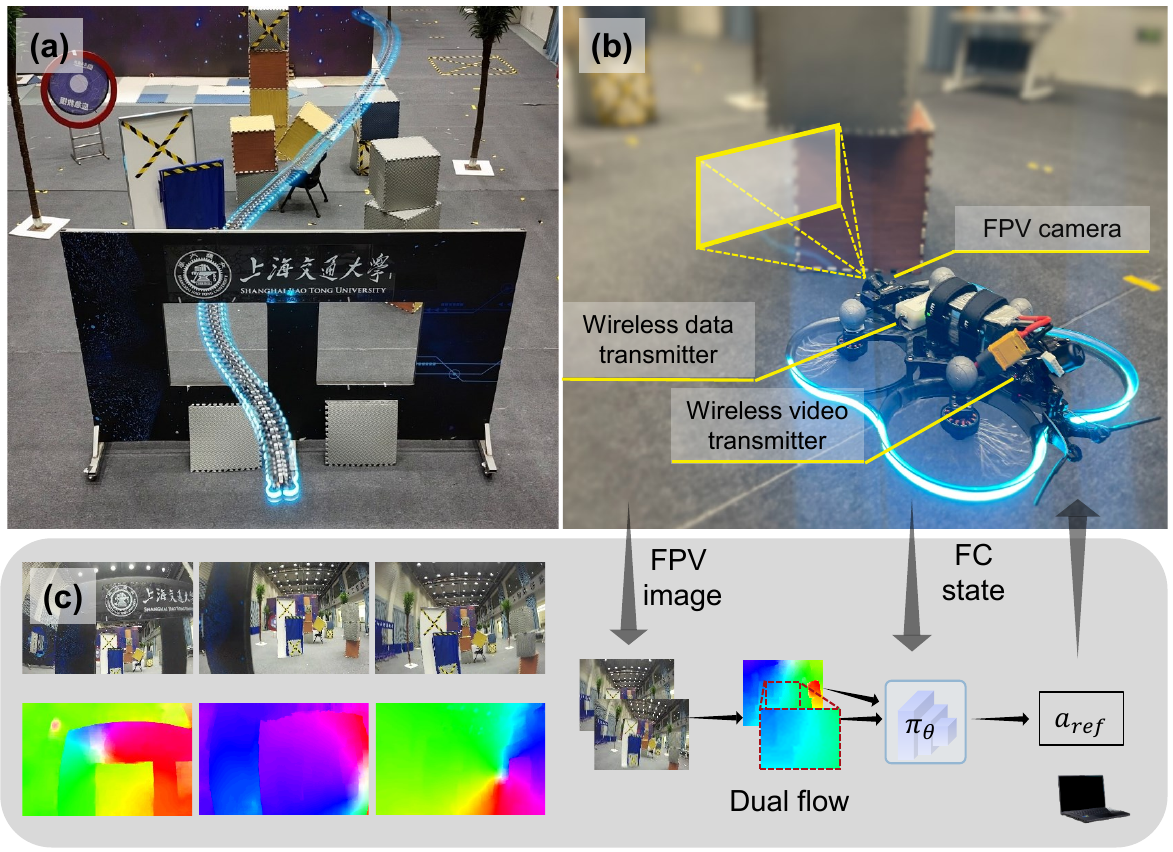}
  \caption{\textbf{Our drone autonomously navigates in the cluttered environment using optical flow estimated from a single camera.} (a) Overview of the testing environment and trajectory. (b) Our real-world offboard control system is equipped with an FPV camera, a wireless video transmitter, and a wireless data transmitter. (c) FPV images and optical flow estimations recorded during the flight. 
  }
  \label{fig:main}
\end{figure} 
 
%

\change{Human FPV pilots demonstrate remarkable abilities to navigate complex environments and perform aggressive maneuvers using only monocular visual feedback. 
%
%
A key research question is how to enable robotic systems to achieve comparable performance with the same input modality.
While continuous efforts~\cite{Loquercio2018DroNet, Gandhi2017Crashing, sadeghi2017cad2} have been made to develop monocular vision-based navigation systems through extracting learnable features from raw image data, these approaches often yield suboptimal performance due to simplified control design or overfitting to specific environments, struggling to improve the agility and generalization. 
}

\change{Optical flow measures the motion of pixels in an image sequence over time, providing an essential connection between pixels and spatial perception.}
Studies have shown that optical flow is fundamental to navigation in flying insects, such as flies and bees, which rely primarily on monocular vision for navigation \cite{egelhaaf2023optic}. Despite having tiny heads and simple neural structures, insects effectively use optical flow for visual odometry \cite{srinivasan1996honeybee}, measuring the gap \cite{ravi2019gap}, tunnel crossing \cite{baird2012visual}, and obstacle avoidance \cite{ravi2022bumblebees}. 
These findings suggest that optical flow provides comprehensive information about self-motion and spatial layouts, showing its potential for autonomous flight. 
Therefore, we ask: \emph{How can we enable autonomous flight using optical flow, without relying on explicit depth measurements?}

 \begin{figure*}[t]
      \centering
      \includegraphics[width=0.97\textwidth]{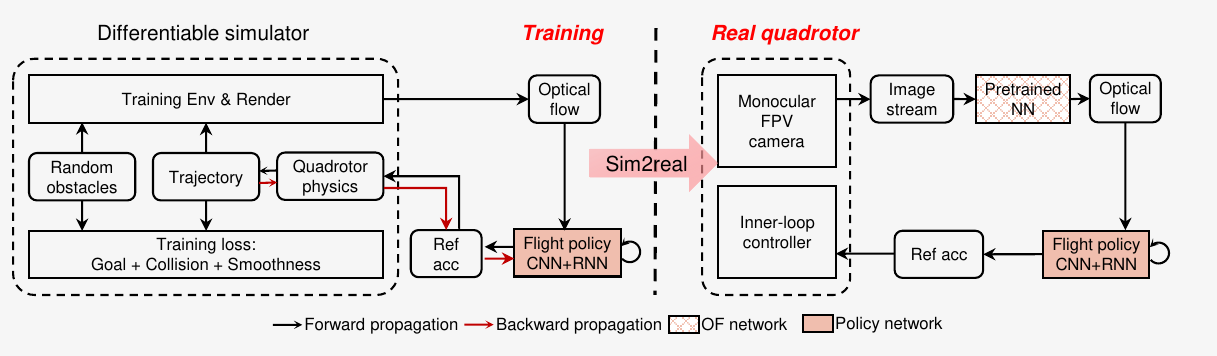}
      \caption{\textbf{System overview}. We train our neural network policy using a differentiable simulator, which enables simulating quadrotor physics, rendering ground-truth optical flow, and calculating analytic policy gradients. We deploy our policy using a real FPV-style quadrotor in the real world. The reference acceleration output by the flight policy is sent into the inner-loop controller.
      }
      \label{fig:text_system}
            \vspace{-0.2em}
\end{figure*} 

Leveraging optical flow for autonomous high-speed navigation in cluttered environments remains a significant challenge. 
First, optical flow estimation is prone to errors, especially in low-texture environments, varying illumination, and high-speed motions. Second, extracting essential motion cues from optical flow, such as time-to-collision~\cite{souhila2007optical}, nearness maps ~\cite{bertrand2015bio}, and looming~\cite{zhao2018bio}, remains an open problem, as existing algorithms rely on simplified assumptions like pure translational motion or front-parallel obstacles. Third, designing effective control strategies based on these 2D cues is difficult due to the lack of intuitive metric information. 

Existing methods usually rely on heuristic strategies like braking~\cite{xiao2021optic}, balancing the flow field~\cite{souhila2007optical}, or simple control approaches, such as PID control~\cite{Croon2021enhancing} or Incremental Non-Linear Dynamic Inversion (INDI)~\cite{HO2024INDI}. While PID controllers are widely used, they often struggle to handle the non-linear dynamics commonly encountered in optical flow-based applications. INDI, on the other hand, is better suited for addressing non-linearities but relies heavily on accurate system modeling and real-time sensor data. 
Consequently, as the vehicle is pushed to higher levels of agility, the limitations of classical control methods become increasingly pronounced, leading to reduced reliability and stability.

\textbf{Contribution:} This work introduces a learning-based approach for flow-based monocular obstacle avoidance in quadrotors. %
\change{
Our method enables a racing drone to navigate through complex, unstructured environments, such as forests, at high speeds (up to~\SI{6}{\meter\per\second}), using only a monocular FPV camera. 
%
%
The key to our approach is a learning framework that tightly integrates flow estimation with an efficient policy optimization method through differentiable simulation. 
First, 
we incorporate central flow attention and action-guided active sensing mechanisms to focus on task-relevant flow information for downstream decision-making. 
%
%
Second, we learn a memory-based policy that maps the predicted optical flow to control commands by first-order gradient optimization through differentiable simulation. 
Third, we develop a GPU-based differentiable simulation that generates optical flow and ground-truth depth for high-speed training, which allows zero-shot transfer of the trained policy to the real world.
Our system is validated using an offboard setup, where video streams from the onboard camera are transmitted to a laptop for real-time processing and control.
An overview of our system is given in Fig.~\ref{fig:text_system}. 
}



\section{Related Work} \label{II}


\subsection{Depth based Flight} 
Traditional approaches to autonomous flight rely on constructing 3D obstacle maps. 
 For instance, mapping-based approaches~\cite{zhou2021raptor, zhou2019fast, voxblox, FIESTA} generate an ESDF (Euclidean Signed Distance Field) map using depth sensors, which is then used for local path search and trajectory optimization.
Other techniques simplify obstacle representation by focusing on obstacle-free spaces~\cite{Bubble, FASTER, tordesillas2020mader}, extracting safe flight corridors, and utilizing the convex hull properties of these corridors to simplify constrained trajectory optimization. 
Despite their successes, the modular pipeline introduces cumulative errors and added latency, which hampers the vehicle's performance, particularly during high-speed flight.

Learning-based methods aim to address these limitations by developing end-to-end planning or control policies. For instance, an imitation learning system~\cite{loquercio_learning_2021} replaces traditional mapping and planning with a neural network planner that converts depth images directly into collision-free trajectories. 
More recent work~\cite{Zhang2024BackTN} employs a first-order gradient optimization technique to train an end-to-end policy that maps depth inputs to commands, enabling agile swarm flight in complex environments.

\subsection{Optical Flow based Flight}

Early research on optical flow flight control primarily focused on adapting mechanisms observed in insects to robotics. However, the complexity of how insects utilize optical flow in vision makes direct adaptation challenging. \change{These bio-inspired approaches often rely on simplified controllers tailored to specific behaviors, with low error tolerance, which restricts their ability to enable agile flight in unknown environments.}
%
For instance, time-to-collision~\cite{nelson1989obstacle}, derived from flow divergence or looming motion cues~\cite{hatsopoulos1995elementary}, is widely employed for collision avoidance. Two common computational methods include determining the Focus of Expansion~\cite{souhila2007optical} and calculating changes in relative size \cite{horn2007time, yang2020upgrading}. However, the former is accurate only during translational motion in static scenes and struggles with moving objects, while the latter is limited to fronto-parallel obstacles. \delete{Additionally, optical flow estimation is prone to errors due to the complexity of the scene and self-motion.} \change{
Recent research on optical flow-based control~\cite{serres2017optic, Croon2021enhancing, Croon_2016, Croonattctrl, HO2024INDI, bouwmeester2023nanoflownet} has made important improvements. 
For instance, de Croon et al.~\cite{Croon2021enhancing} introduces a flow-based controller that enables UAVs to perform landing and low-speed obstacle avoidance by first recovering depth through self-induced oscillations and then applying a proportional gain or stop-and-turn control strategy.
NanoFlownet~\cite{bouwmeester2023nanoflownet} combines a lightweight neural network for real-time optical flow prediction with simple PD control, allowing tiny drones with minimal sensors to avoid obstacles. 
However, existing methods have primarily demonstrated low-speed flight in simplified scenarios. In contrast, a key contribution of our work is demonstrating that end-to-end approaches can enable high-agility flow-based flight in unknown, cluttered, and natural environments.
}

\section{Optical flow as visual cues}
\label{sec:visual_cues}

\begin{figure}[htbp] 
      \centering
      \includegraphics[width=0.48\textwidth]{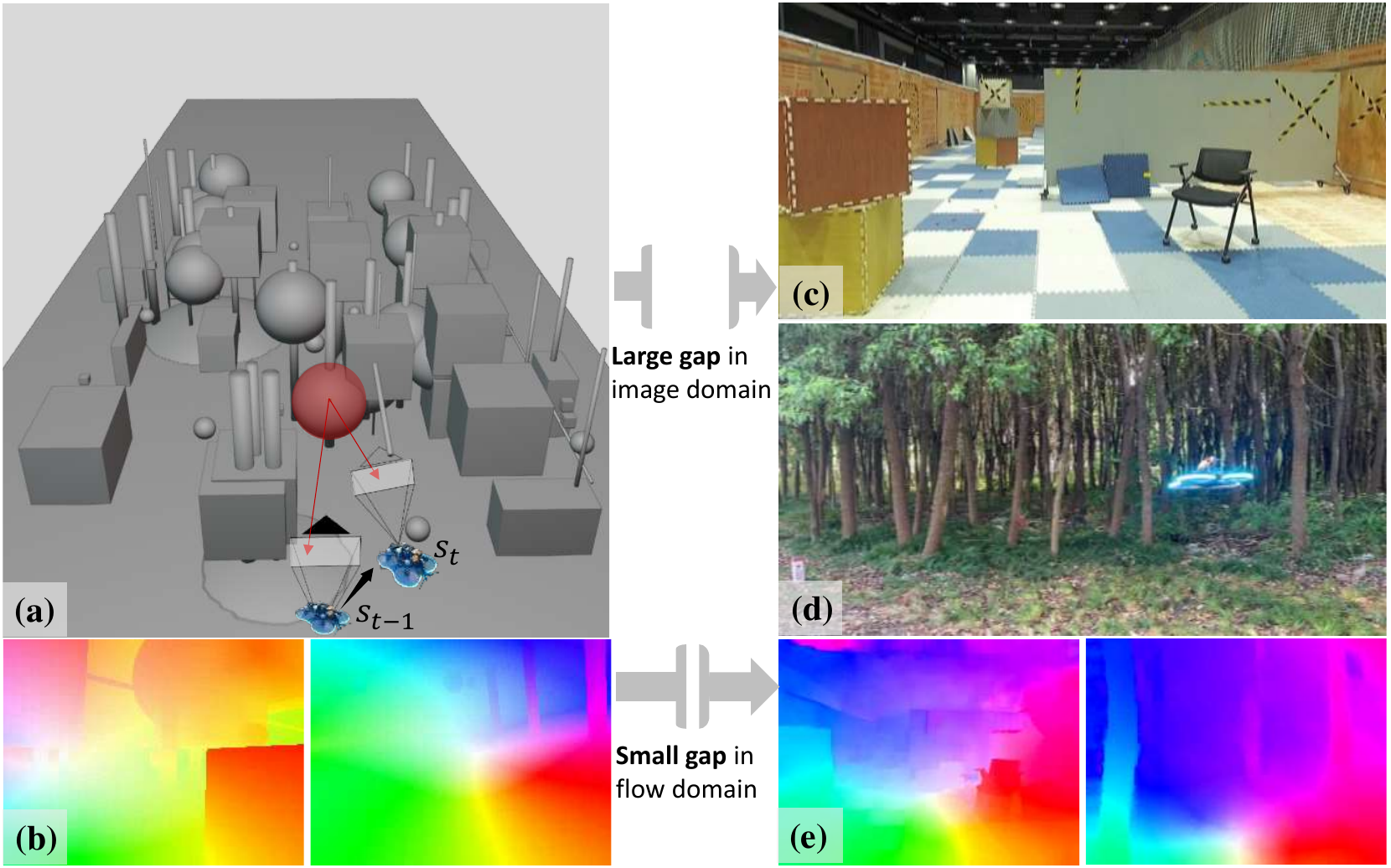}
      \caption{
      \textbf{Our training environment vs Real-world testing environments.} (a) Our training environment features objects with simple geometric shapes. (b) Sampled ground-truth optical flow for training. (c-d) Real-world testing environments. (e) Estimated optical flow in the real world. Flow-based representation captures essential information about motion and removes redundant information that might be irrelevant for obstacle avoidance. 
      }
      \label{fig:domain_gap}
\end{figure}

Our goal is to achieve autonomous obstacle avoidance using a monocular camera that captures RGB images. While it is possible to directly use raw RGB images for learning navigation policies, employing optical flow as the visual input offers two main advantages.
First, optical flow captures the relative 3D motion with respect to the environment, eliminating redundant information present in the RGB domain. As illustrated in Fig.~\ref{fig:domain_gap}, although the visual appearance (RGB images) in our training environment is very different from that of the actual deployment environments, these differences can be minimized in the flow domain. This small domain gap ensures the policy's ability to transfer effectively from simulation to reality.
Second, optical flow can be rendered in parallel at high speed by leveraging a simple geometric simulator, as shown in Fig.~\ref{fig:domain_gap}~(a). This approach is more efficient and simpler than photorealistic RGB simulation, ensuring a time-efficient training process.

We briefly review the relationship between depth, optical flow, and camera motion using the following equation:
\begin{equation}\label{eq:flow_depth_spd}
\small
    \begin{aligned}
        \left[\begin{array}{c}
            \dot{\overline{p}}_{x} \\
            \dot{\overline{p}}_{y}
        \end{array}\right]
        &= \frac{1}{p_{z_{b}}} \left[\begin{array}{ccc}
            -1 & 0 & \overline{p}_{x} \\
            0 & -1 & \overline{p}_{y}
        \end{array}\right] {}^{b}\mathbf{v} \\
        &\quad + \left[\begin{array}{ccc}
            \overline{p}_{x}\overline{p}_{y} & -(1+\overline{p}_{x}^{2}) & \overline{p}_{y} \\
            1+\overline{p}_{y}^{2} & -\overline{p}_{x}\overline{p}_{y} & -\overline{p}_{x}
        \end{array}\right]
        {}^{b}\mathbf{\omega}.
    \end{aligned}
\end{equation}
Here, $\overline{p}_{x}$, $\overline{p}_y$ are the x and y displacement in image plane with unit of \SI{}{m}. ${}^{b}\mathbf{v}$, ${}^{b}{\omega}$ are the linear velocity and angular velocity in the body frame. 
We formulate the flow using a pinhole camera model, assuming that the drone's body frame coincides with the camera frame, where the forward direction aligns with the camera frame's z-axis. The first part of the right-hand side can be referred to as the translational optical flow, as it is related to the ratio of velocity and depth. The second term on the right-hand side can be termed the rotational optical flow since it is solely associated with rotation.



\begin{figure}[htbp] 
      \centering
      \includegraphics[width=0.48\textwidth]{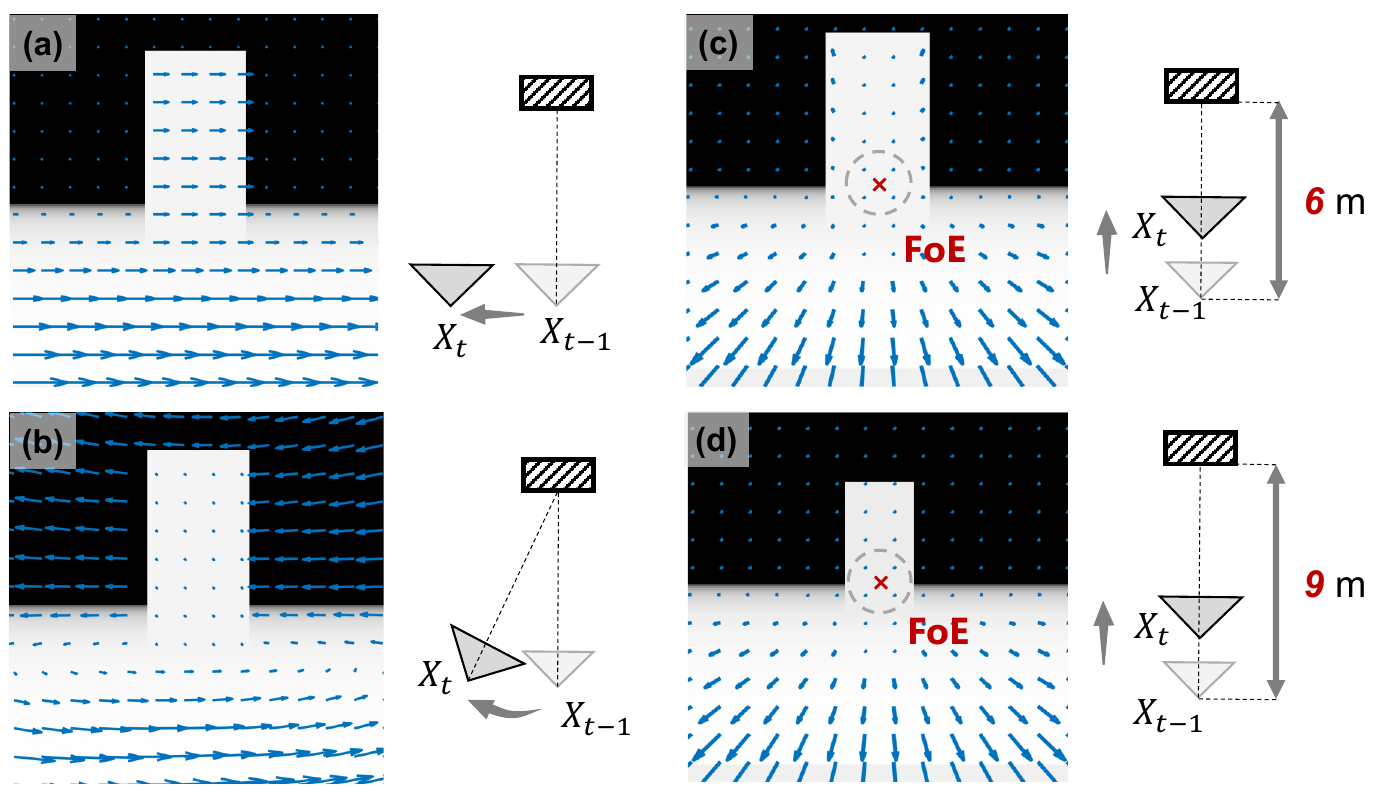}
      \caption{
      \textbf{Challenges in detecting obstacles from optical flow}.
      \textbf{Left:} During rotation (b), the flow of obstacles can vanish, making it appear similar to the background flow in the pure translation scenario (a).  
      \textbf{Right:} Flow values near the Focus of Expansion (FoE) are minimal, making it difficult to detect looming effects using local flow divergence with noisy estimation.
      }
      \label{fig:flow_ambi}
\end{figure}

The equation indicates that optical flow generated by translational motion encodes depth information about obstacles, increasing in magnitude as the drone approaches them. Therefore, a straightforward way to detect obstacles is by monitoring the flow magnitudes. However, this approach becomes unreliable when rotational motion is involved, as the flow caused by rotation contains no depth information. As shown in Fig.~\ref{fig:flow_ambi} (a) and (b), 
mistaking rotational flow for the evidence of approaching obstacles can result in false actions. \change{This occurs because camera rotation generates optical flow similar to that caused by obstacle proximity, which obscures accurate depth perception. As a result, constructing reliable distance-based 3D maps from optical flow becomes challenging for traditional map-based planning methods.} 
Even if we manage to exclude the flow caused by rotation, the flow is less informative at the velocity projection point on image plane, also known as FoE (Focus of Expansion) where $\begin{bmatrix}-1 & 0 & \overline{p}_{x}\end{bmatrix}{}^{b}\mathbf{v}=0$ and  $\begin{bmatrix}0 & -1 & \overline{p}_{y}\end{bmatrix}{}^{b}\mathbf{v}=0$. 
The depth information ${p_{z_{b}}}$ is unobservable by optical flow at FoE, potentially causing insensitivity in the critical flight direction. Fig.~\ref{fig:flow_ambi} (c) and (d) show that in scenarios where \SI{2}{\meter}-wide obstacle is positioned at distances of \SI{6}{\meter} and \SI{9}{\meter}, the optical flow near FoE exhibits similarity with small value, causing sensitivity to noise. 

Therefore, utilizing optical flow for obstacle avoidance presents significant challenges, particularly due to the inherent complexities of dense flow estimation from pixel motions and the additional difficulties posed by flow ambiguity. These factors make quadrotor trajectory planning and control in the optical flow domain highly challenging.




\section{Method}\label{method}

\subsection{Problem Formulation}
We formulate the quadrotor obstacle avoidance task as an optimization problem. The robot is modeled via a discrete-time dynamical system \( x_{k+1} = f(x_k, u_k) \), characterized by continuous state and control input spaces, denoted as \( \mathcal{X}  \) and \( \mathcal{U} \), respectively. At each time step \( k \), the system state is \( x_k \in \mathcal{X} \), and the corresponding control input is \( u_k \in \mathcal{U} \).
We represent the quadrotor's dynamics using a differentiable point-mass model with acceleration as the control input.
%
%
The vehicle receives a cost signal \(l_k(x_k, u_k)\) at each time step, which is used to evaluate the state and action. 
Additionally, the robot observes the environment through an RGB image \(o_k\) received from a monocular camera mounted the vehicle. 

We represent the control policy as a neural network~\( u_k = \pi_{\theta} (OF_k)\), which takes an optical flow \( OF_k \) as input and outputs the control input \( u_k \).
The objective of our optimization problem is to find the optimal policy parameters \(\theta^{\ast}\) by minimizing the total loss via gradient descent:
\begin{align}
\small
    \min_{\theta} \mathcal{L}_{\theta} &=  
      \sum_{k=0}^{N-1} l(x_{k}, u_k) 
    = 
      \sum_{k=0}^{N-1} l(x_{k}, \pi_{\theta}(OF_k))
      \\
    \theta & \leftarrow \theta - \gamma \nabla_{\theta} \mathcal{L}_\theta
\end{align}
where $\gamma$ is the learning rate.
For policy training, we simulate ground-truth optical flow using our rendering engine.
Given that we can only perceive the environment via an RGB camera,
we use the NeuFlow~\cite{zhang2024neuflow} to estimate the optical flow $\hat{OF}_k = \text{NeuFlow}(o_k, o_{k-1})$ for real-world deployment. NeuFlow uses two consecutive images $o_k$ and $o_{k-1}$ captured by the camera.  
It is a state-of-the-art neural network architecture that achieves high-accuracy flow prediction and efficient inference on edge computing platforms. 

\subsection{Rendering Optical Flow}
To render ground-truth optical flow for policy training, we implement a rendering engine. 
Given the current state, including position and orientation, of the vehicle, a ground-truth depth image is generated using ray tracing, where the 3D geometry of obstacles in the scene is solved to identify the intersection points along the rays. The nearest intersection point is then selected to determine the depth value for each pixel.
To compute the optical flow at each time step ${OF_k}$, we use the depth image from the previous time step ${Depth_{k-1}}$ as a reference frame $C_{k-1}$. Each pixel $(u_{k-1}, v_{k-1})$ with its associated depth value ${Depth_{k-1}}$ is converted into a 3D point cloud. This point cloud is then reprojected onto the current image plane $C_k$, based on the drone's state, to obtain the new pixel positions $(u_k, v_k)$. The ground truth optical flow ${OF_k}$ is computed as ${OF_k} = (u_k - u_{k-1}, v_k - v_{k-1})_{2 \times h \times w}$.
With our CUDA~\cite{cuda} parallelized implementation, our simulator achieves over \SI{350000}{frames\per\second} at a resolution of $48 \times 64$ on a single Nvidia RTX 3090 GPU. 
We utilize ultra-low resolution optical flow \(12 \times 16\) as input to the policy network for two key reasons: 1) low-resolution flow enables more efficient training and deployment, and 2) downscaling a dense optical flow acts as a low-pass filter, preventing the policy from overfitting to irrelevant details.

\subsection{Central Flow Attention with Active Sensing}
While low-resolution optical flow offers several benefits, it might sacrifice important details. To address this, we introduce a central flow attention mechanism. Specifically, the central region of the original high-resolution optical flow is selected and downscaled; then, it is concatenated with the downscaled full flow to form a dual input for the control policy.

\begin{figure}[h] 
      \centering
      \includegraphics[width=0.48\textwidth]{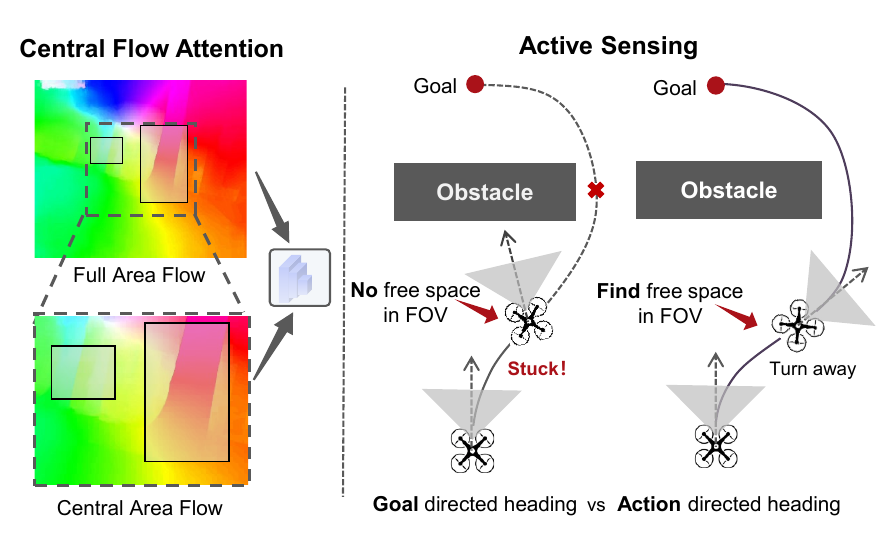}
      \caption{
      We combine a central flow attention mechanism (left) with active sensing (right) to extract task-relevant flow information while maintaining efficient computation. 
      }
      \label{fig:activeyaw}
\end{figure} 
This concept is inspired by human pilots, who use eye gaze to concentrate on task-relevant visual information, e.g., the target gate, during drone racing~\cite{pfeiffer2021human}. This dual-resolution flow approach improves inference time by representing the optical flow of the entire image with a more compact matrix; at the same time, it enhances focus on the image center, enabling the network to preserve crucial details in this region, leading to more precise decision-making. 
%

\change{For effective obstacle avoidance, we also combine flow estimation with active robot sensing. Specifically, central flow attention retains vital obstacle information in the image center. However, the effectiveness of central flow attention heavily depends on the alignment between the image center and the quadrotor's motion direction. To address this challenge, we leverage the quadrotor's differential flatness property, specifically its independent yaw control capability, to actively align the camera with the flight direction. This active perception strategy ensures that the most critical obstacle information is consistently captured in the high-resolution central region. By maintaining this optimal alignment between sensing and motion, our system not only enhances the effectiveness of dual-resolution flow processing but also enables efficient free space exploration during obstacle avoidance maneuvers, as shown in Fig.~\ref{fig:activeyaw}.}

\subsection{Policy Optimization via Differentiable Simulation}
\label{subsec:diffsim}
\begin{figure}[h]
\vspace{-1.0em}
      \centering
    \includegraphics[width=0.44\textwidth]{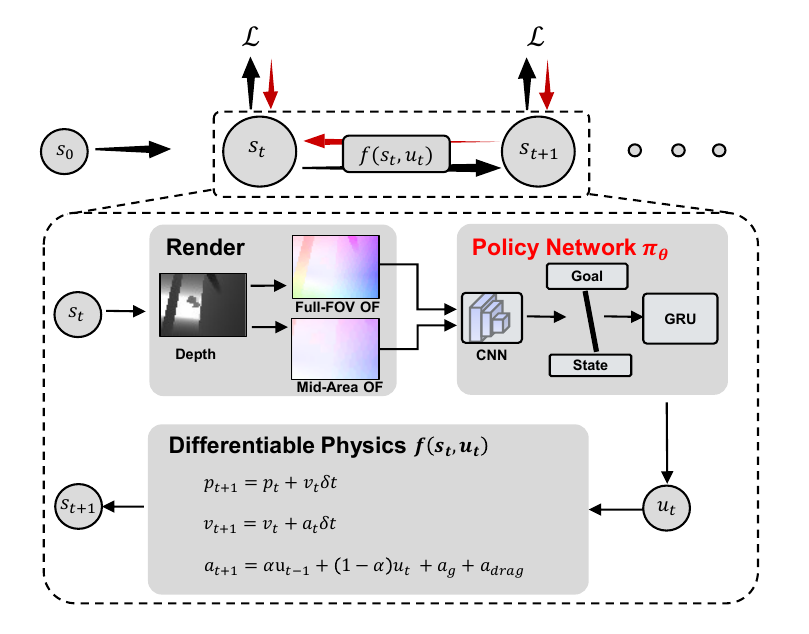}
      \caption{\textbf{Overview of our differentiable simulator.} We use a point mass differentiable model for forward simulation (black arrow) and gradient calculation (red arrow). The controller response parameter is calibrated as $\alpha=e^{-\lambda (t - \tau)}$ where $\tau={2\over{15}}$ and $\lambda=15$ in our platform. }
      \label{fig:pipeline}
\end{figure}

To optimize the control policy, we developed a high-performance simulator capable of simulating a point-mass model~\cite{Zhang2024BackTN}, a simplified yet effective representation of quadrotor dynamics, and rendering ground-truth depth and optical flow (Fig.~\ref{fig:pipeline}). 
An important feature of our simulator is its ability to construct a computational graph that enables \emph{automatic differentiation}, crucial for gradient-based policy optimization methods such as backpropagation, which is widely used for training neural networks.
During the forward simulation, the vehicle’s trajectory is generated by randomly initializing the state of the vehicle and repeatedly applying the control commands output by the policy. Once the simulation completes, the computational graph is built. This allows us to apply Backpropagation Through Time (BPTT) to update the policy parameters efficiently by calculating gradients over the entire trajectory. The policy gradient can be expressed as 
$\partial \mathcal{L}_{\theta} / \partial \theta = $
\begin{equation}
  \small
       \frac{1}{N}\sum_{k=0}^{N-1} \left( \sum_{i=0}^{k} \frac{\partial l_k }{\partial x_{k}} \prod_{j=i+1}^{k} \left( \frac{\partial x_{j}}{\partial x_{j-1}} e^{-\alpha \Delta t} \right) \frac{\partial x_i}{\partial \theta} + \frac{ \partial l_k}{\partial u_k} \frac{ \partial u_k}{\partial \theta} \right) 
      .
\end{equation}
where $\alpha$ is an exponential decay of the dynamics gradients.

\subsection{Loss Function}
We define a navigation task where the objective is to follow a target velocity while avoiding obstacles similar to ~\cite{Zhang2024BackTN}.
We formulate a loss function that has four terms: velocity tracking $\mathcal{L}_v$, obstacle avoidance $\mathcal{L}_c$, and control smoothness $\mathcal{L}_a$ and $\mathcal{L}_j$.
Specifically, the velocity tracking loss is defined as 
\begin{equation}
\small
    \mathcal{L}_v=\frac{1}{T} \sum_{k=1}^T \text{Smooth L1}
    (\|v_k^{ref}-\bar{v}_k \|_2,0)
\end{equation}
where $v_k^{ref}$ is the reference velocity and $\bar{v}_k$ is the average velocity calculated via a moving average. 
We use an averaged speed instead of current speed to remove potential high-frequency oscillations, which has shown empirically better performance. 

We define a collision avoidance loss as the following
\begin{equation*}
\small
  \mathcal{L}_c=\frac{1}{T} \sum_{k=1}^T v_k^c \max(1-(d_k-r_q),0)^2+\beta_1 \ln(1+e^{\beta_2 (d_k-r_q)}),
\end{equation*}
where $d_k$ represents the closest distance to the obstacle, $r_q$ is the radius of the quadrotor, and $v_k^c$ is the speed component directed towards the obstacle. 
The term $v_k^c$ functions as an adaptive weight, dynamically adjusting the importance of proximity-based collision risk. If the agent is moving away from an obstacle, even when the distance is small, the risk is minimal. On the other hand, if the agent is approaching an obstacle at high speed, even when the obstacle is relative far, the risk can increase. The second term provides a soft and smooth penalty that ensures more robust obstacle avoidance, particularly in scenarios where obstacles are not immediately close but still pose a risk. Here, $\beta_1$ and $\beta_2$ are hyperparameters.

To avoid large oscillation in the action space, we define two smoothness losses to penalize the acceleration and jerk:
\begin{equation*}
\small
  \mathcal{L}_a=\frac{1}{T} \sum_{k=1}^T \|a_k \|^2,
  \qquad
  \mathcal{L}_j=\frac{1}{T-1} \sum_{k=1}^{T-1} \|\frac{a_k-a_{k+1}}{\Delta t}\|^2.
\end{equation*}
These losses help maintain a smoother flight, improving the quality of optical flow estimation and, hence, obstacle avoidance performance. The final loss is the weighted sum of these four terms, $L=\lambda_v \mathcal{L}_v+\lambda_c \mathcal{L}_c+\lambda_a \mathcal{L}_a+\lambda_j \mathcal{L}_j$, \change{where a coordinate descent optimization approach is used to determine the weights. In out implementation, we set} $\lambda_v=1, \lambda_c=2, \lambda_a=0.015, \lambda_j=0.003$ .

\begin{figure*}[th]
      \centering
      \includegraphics[width=0.95\textwidth]{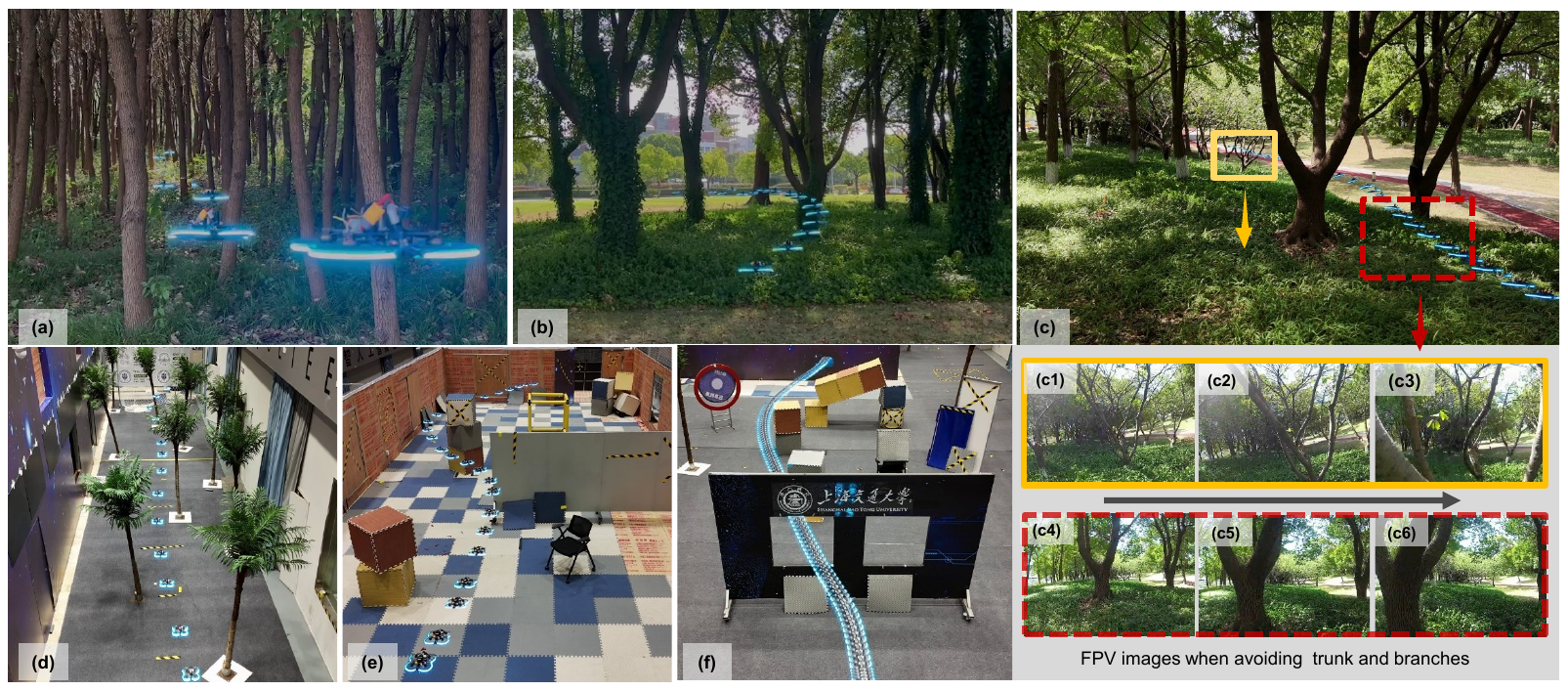}
      \caption{\textbf{Real world experiment results.} We validate our algorithm in various unknown environments, including (a) dense forest, (b) sparse forest, (c) trees with branches, and cluttered indoor scenes (e) and (f). The quadrotor's flight trajectory is visualized by overlaying its position in the video footage (please refer to the supplementary video for more details).}
      \label{fig:real_result}
         \vspace{-1.5em}
\end{figure*}

\section{Experiments}


\subsection{Experiment Setup}
We train our neural network control policy using our CUDA-based differentiable simulator (described in Sec.~\ref{subsec:diffsim}).
Training on a standard computer equipped with an Intel i9 CPU and an Nvidia RTX 3090 GPU, the policy converges in approximately 2 hours.
%

\subsection{Policy Training Performance}
\begin{figure}[htbp] 
  \centering
  \includegraphics[width=0.48\textwidth]{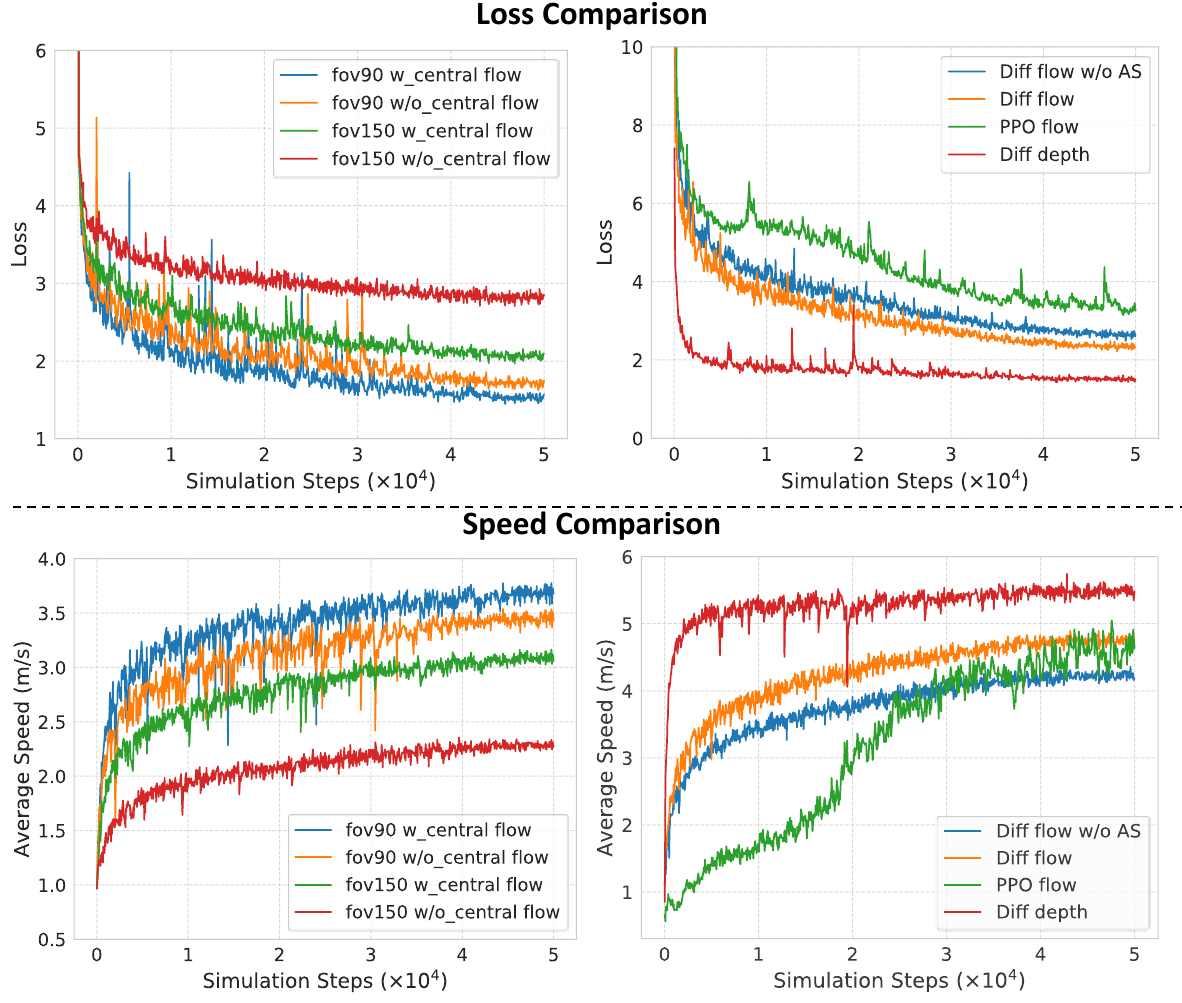}
  \caption{\textbf{Training performance.} The first column demonstrates the effectiveness of Central Flow Attention by comparing policies with or w/o the central flow under different camera FOV settings. The policy with Central Flow Attention mechanism has higher average speed and lower total loss. The second column includes ablation study on input modality (depth vs. flow), yaw strategy (active sensing (AS) vs. w/o AS), training method (PPO~\cite{schulman2017proximalpolicyoptimizationalgorithms} vs. differentiable physics (Diff)). The reference speed varies from \SI{1.5}{\meter\per\second} to \SI{12}{\meter\per\second}. }
   \vspace{-0.8em}
  \label{fig:Central_flow_abl}
\end{figure}

%
We demonstrate the effectiveness of our approach for policy training using differentiable simulation and optical flow in obstacle avoidance. 
We show that the introduced Central Flow Attention mechanism \change{and active sensing} can significantly improve the convergence speed and asymptotic performance of the policy, shown in~Fig.~\ref{fig:Central_flow_abl}. 
With a smaller field of view (FOV), e.g., 90$^\circ$, the flow naturally contains enough detailed information about obstacles in the environment; the central flow of attention allows marginal improvement.  
When using a large FOV, e.g., 150$^\circ$, the captured image provides a broader environmental context but lack fine detail in specific regions, such as the flight path. 
In this case, the central optical flow offers crucial compensatory information, directing the control policy’s attention toward the most relevant features for navigation, and hence, improves the policy performance significantly.
\change{Additionally, the right column in Fig.~\ref{fig:Central_flow_abl} illustrates that aligning yaw with the direction of motion outperforms alignment with the goal position. We hypothesize that this action-guided active sensing strategy enhances the impact of the central flow.}

%
%
%

We compare the performance of flow-based and depth-based policy training to highlight that learning to control from optical flow is a challenging task by definition. 
The learning curves are shown in Fig.~\ref{fig:Central_flow_abl}.  
%
Depth image directly encodes the distance between the vehicle and obstacles, providing clear and actionable guidance for navigation. In contrast, optical flow offers more ambiguous information (illustrated via Fig.~\ref{fig:flow_ambi}) as it represents motion rather than explicit distance.
As a result, given only flow observations, a representation of the environment needs to be learned as part of learning control policies, making flow-based obstacle avoidance more challenging.

\change{We further compare the policy trained by using Proximal Policy Optimization (PPO)~\cite{schulman2017proximalpolicyoptimizationalgorithms}. 
The results show that the differentiable method outperforms PPO in terms of both convergence speed and final loss. In optical flow-based obstacle avoidance, the observation of optical flow and the action are tightly coupled. PPO, requiring action sampling, can introduce fluctuations in flow observations, leading to biased distribution and slower convergence. In contrast, the differentiable method can provide first-order analytical gradients, leading to faster convergence with lower losses.}

\subsection{Sim-to-Sim : A Baseline Comparison}
\begin{figure}[ht] 
  \centering
  \includegraphics[width=0.48\textwidth]{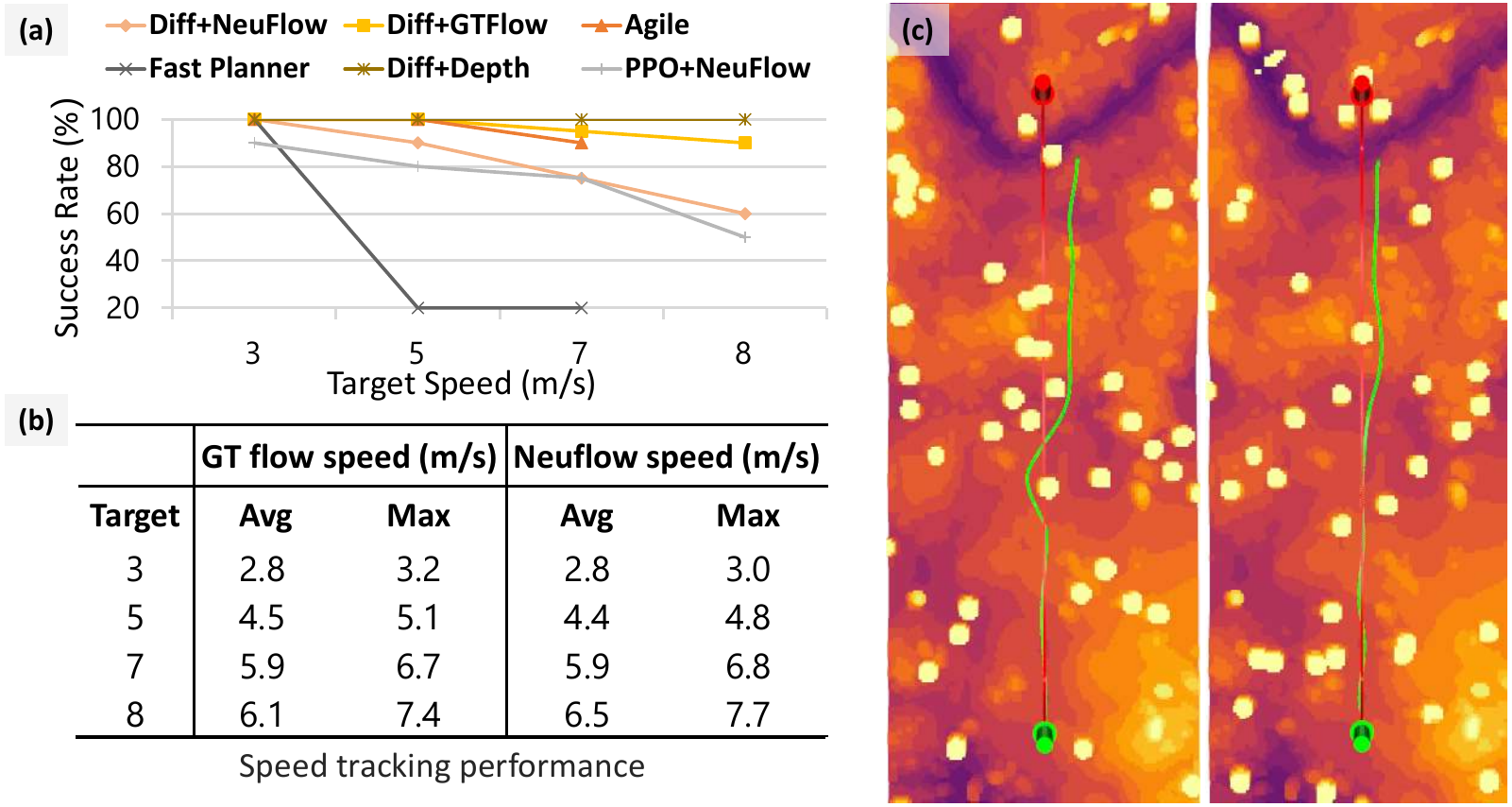}
  \caption{\textbf{Baseline comparisons for sim-to-sim transfer.} We conducted 20 runs for each method. (a) success rates comparison of our algorithm and other SOTA methods in the same simulator with different target speeds. (b)  velocity tracking performance. (c) visualization of the drone's trajectories and the testing environments.}
   \vspace{-0.8em}
  \label{fig:sim_result}
\end{figure}

This section studies the generalization capability of our flow-based policy in an unseen environment. Additionally, we compare the performance of our policy against state-of-the-art depth-based and mapping-based methods~\cite{loquercio_learning_2021, zhou2019fast}.
We use a forest environment proposed in~\cite{loquercio_learning_2021} and the Flightmare simulator~\cite{song2020flightmare} for evaluation. 
A top-down view of the forest environment is given in~Fig.~\ref{fig:sim_result}~(c). 
The results are shown in Fig.~\ref{fig:sim_result}~(a), our flow-based policy successfully generalizes to the unseen forest environment without fine-tunning and achieves a high success rate at high speed, e.g., at~\SI{6}{\meter\per\second} compared to state-of-the-art depth-based learning methods. Notably, when using ground-truth optical flow, our policy achieved a 90\% success rate even at speeds of~\SI{8}{\meter\per\second}. However, when utilizing the real-time lightweight optical flow network, NeuFlow, the success rate dropped to 60\%. 
This decline is attributed to the challenges posed by large displacement in optical flow during fast movement, which increases the error in optical flow estimation. 
These findings suggest that the current limitations in dense optical flow accuracy remain a constraint on our approach's overall performance.
Fig~\ref{fig:sim_result}~(b) illustrates the velocity tracking performance, demonstrating that the policy is able to closely track and achieve velocities near the target speed while avoiding obstacles. 

\subsection{Sim-to-Real} 
We conduct extensive real-world experiments to demonstrate the strong generalization capability of our policy. Specifically, we test in a variety of environments (shown in Fig~\ref{fig:real_result}), including dense forests and cluttered indoor scenarios, featuring diverse obstacles such as thin branches, narrow gaps, walls, and chairs. Despite training our policy solely in a simple 3D environment with a point mass model, the policy transfers directly to these complex real-world settings in a zero-shot manner, achieving a maximum speed of \SI{6}{\meter\per\second}. To the best of our knowledge, this represents one of the most agile flights achieved in complex real-world conditions using an RGB camera combined with optical flow estimation. 

The strong generalization and agile flight performance of our system are attributed to both the accurate perception network and the robust control policy. 
The perception network (NeuFlow) provides real-time, relatively accurate optical flow predictions from captured images.
Additionally, we combine these predictions with our proposed central flow attention mechanism and active sensing, allowing the robot to focus on its flight path while maintaining awareness of the broader environment.
Furthermore, policy optimization using differentiable simulation enables end-to-end optimization of a control policy that maps optical flow directly to control commands via gradient-based optimization. This approach eliminates the need for human intuition, unlike traditional rule-based or heuristic controllers, leading to more robust performance, particularly at high speeds where flow estimation becomes challenging due to large motion in the image.

\begin{figure}[h] 
      \centering
    \includegraphics[width=0.48\textwidth]{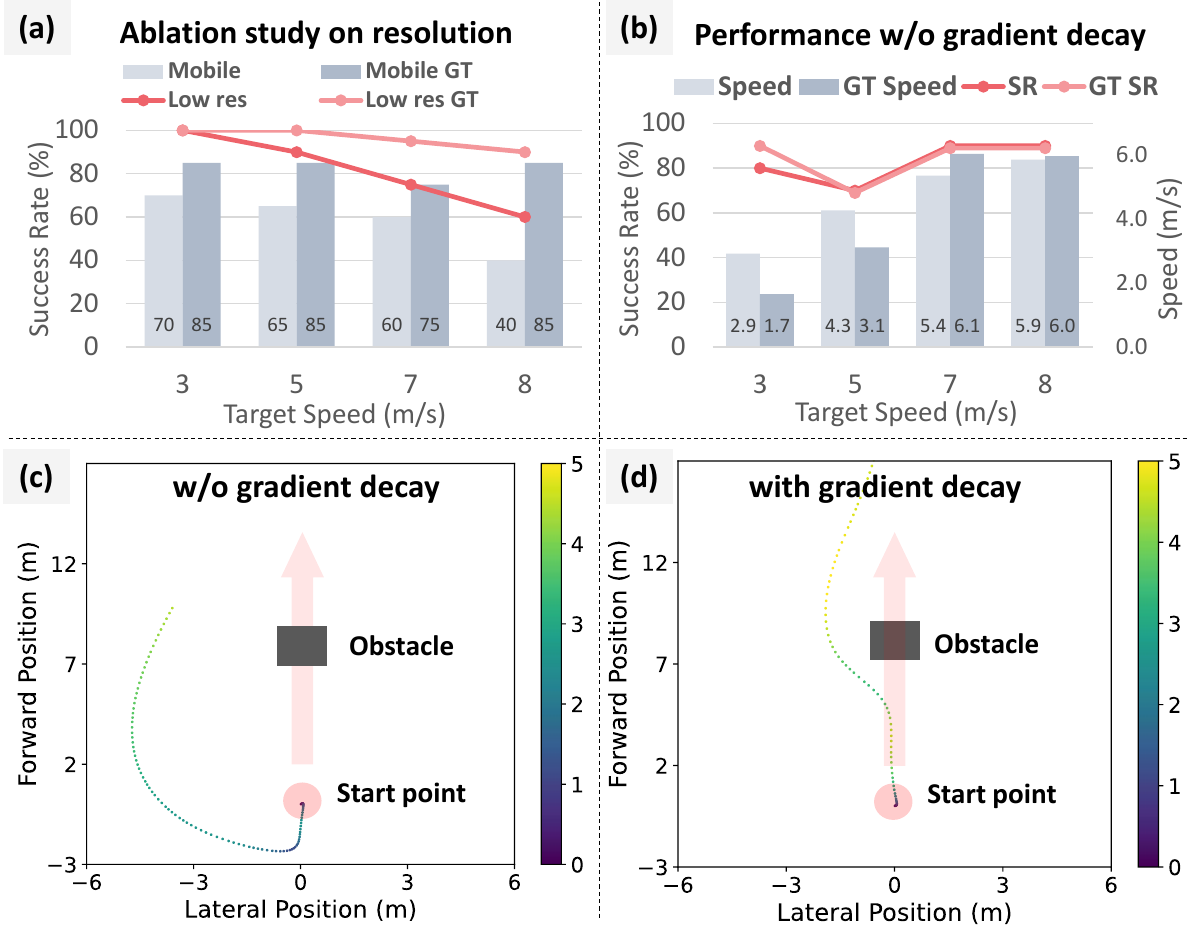}
      \caption{\textbf{Ablation studies}: flow resolution and gradient decay.}
       \vspace{-0.8em}
      \label{fig:abl}
\end{figure} 

\subsection{Additional Studies} 
\label{subsec:discussion}
This section provides additional studies on the impact of optical flow resolution and the use of gradient decay for policy optimization, shown in Fig~\ref{fig:abl}. 

\textbf{High-resolution vs Low-resolution Optical Flow:}
The default resolution of the optical flow for control is $12\times16$. We experimented with a higher resolution of $96\times128$ using the MobileNet V3 architecture~\cite{mobilenetv3}. 
As shown in Fig.~\ref{fig:abl}, increasing the optical flow resolution and using a deeper network did not yield significant improvements in obstacle avoidance performance.
Interestingly, when ground-truth optical flow was used, MobileNet achieved a higher success rate, but it still performed worse than the lower-resolution input. This suggests that a smaller policy network with coarser interpolation offers greater robustness to errors introduced by flow estimation. Although ground-truth flow combined with a deeper network can encode more precise obstacle information, especially for small obstacles, the policy network tends to overfit the training environment, making it less robust to errors and diversity introduced by the optical flow estimation during real-world deployment.

\textbf{Gradient Decay:}
Regarding the effectiveness of temporal gradient decay we use in policy gradient, we conducted tests in the simulation, with results as shown in the Fig~\ref{fig:abl}. RNNs without gradient decay propagate gradients back through time. This leads to gradients being backpropagated to early actions where obstacles that have not yet been observed, causing the agent's behavior to be overly conservative and even retreat from the initial state. Although the UAVs demonstrated a high success rate, their maximum speed significantly deviated from the desired speed. We treat this policy as not applicable in the real world. The function of gradient decay is to facilitate the appropriate propagation of gradients to adjacent frames, ensuring that our policy avoids obstacles within its observed range. 

\section{Conclusion}
This work introduces a learning-based method that achieves one of the most agile flow-based flights in natural environments using a monocular camera. 
Our real-world demonstrations indicate that optical flow is an effective representation for obstacle avoidance as it encodes valuable information about the vehicle motion and the surrounding environment.
%
%

%

\par{\textbf{Limitations:}}
\change{While our work demonstrates agile flight with optical flow, a performance gap persists compared to depth-based navigation. This gap arises from optical flow's ambiguity in representing obstacles, particularly near the Focus of Expansion (FOE) or with significant rotational flow components as we have discussed in Sec. \ref{sec:visual_cues}. Our policy also encounters failure cases in these scenarios at high speeds. Future work should focus on improving flow estimation accuracy near the FOE and mitigating the impact of rotational components to enhance obstacle awareness.} 

\balance

\bibliographystyle{unsrt}
\bibliography{IEEEexample}



\end{document}